\documentclass[final]{cvpr}

\usepackage{times}
\usepackage{epsfig}
\usepackage{graphicx}
\usepackage{amsmath}
\usepackage{wrapfig}
\usepackage{amssymb}
\usepackage{multirow}
\usepackage{pifont}
\newcommand{\cmark}{\ding{51}}%
\newcommand{\xmark}{\ding{55}}%
\usepackage{array}
\usepackage{xcolor}
\usepackage{enumitem}
\usepackage[normalem]{ulem}
\usepackage[linesnumbered,ruled]{algorithm2e}
\usepackage[title]{appendix}
\newcommand{\PreserveBackslash}[1]{\let\temp=\\#1\let\\=\temp}

\newcolumntype{C}[1]{>{\PreserveBackslash\centering}p{#1}}
\newcolumntype{R}[1]{>{\PreserveBackslash\raggedleft}p{#1}}
\newcolumntype{L}[1]{>{\PreserveBackslash\raggedright}p{#1}}

\usepackage[pagebackref=true,breaklinks=true,letterpaper=true,colorlinks,bookmarks=false]{hyperref}

\def\cvprPaperID{2546} % *** Enter the CVPR Paper ID here
\def\confYear{CVPR 2021}

\begin{document}

\setlength{\abovedisplayskip}{3pt}
\setlength{\belowdisplayskip}{3pt}

%%%%%%%%% TITLE
\title{Weakly Supervised Instance Segmentation for Videos \\with Temporal Mask Consistency}

\author{Qing Liu\thanks{This work is done during Qing Liu's internship at Facebook.}\\
Johns Hopkins University\\
{\tt\small qingliu@jhu.edu}
\and
Vignesh Ramanathan\\
Facebook\\
{\tt\small vigneshr@fb.com}
\and
Dhruv Mahajan\\
Facebook\\
{\tt\small dhruvm@fb.com}
\and
Alan Yuille\\
Johns Hopkins University\\
{\tt\small alan.l.yuille@gmail.com}
\and
Zhenheng Yang\\
Facebook\\
{\tt\small zhenheny@gmail.com}
}

\maketitle
%\thispagestyle{empty}

%%%%%%%%% ABSTRACT
\begin{abstract}
\vspace{-0.04\linewidth}
    Weakly supervised instance segmentation reduces the cost of annotations required to train models. However, existing approaches which rely only on image-level class labels predominantly suffer from errors due to (a) partial segmentation of objects and (b) missing object predictions. We show that these issues can be better addressed by training with weakly labeled videos instead of images. In videos, motion and temporal consistency of predictions across frames provide complementary signals which can help segmentation. We are the first to explore the use of these video signals to tackle weakly supervised instance segmentation. We propose two ways to leverage this information in our model. First, we adapt inter-pixel relation network (IRN)~\cite{ahn2019weakly} to effectively incorporate motion information during training. Second, we introduce a new \textit{MaskConsist} module, which addresses the problem of missing object instances by transferring stable predictions between neighboring frames during training. We demonstrate that both approaches together improve the instance segmentation metric $AP_{50}$ on video frames of two datasets: Youtube-VIS and Cityscapes by $5\%$ and $3\%$ respectively.
\end{abstract}
\vspace{-0.06\linewidth}
%%%%%%%%% BODY TEXT
\section{Introduction}
\vspace{-0.02\linewidth}
Instance segmentation is a challenging task, where all object instances in an image have to be detected and segmented. This task has seen rapid progress in recent years~\cite{he2017mask,liu2018path,chen2019hybrid}, partly due to the availability of large datasets like COCO~\cite{lin2014microsoft}. However, it can be forbiddingly expensive to build datasets at this scale for a new domain of images or videos, since segmentation boundaries have to be annotated for every object in an image.

\begin{figure}[t]
\vspace{-0.02\linewidth}
\centering
\includegraphics[width=0.93\linewidth]{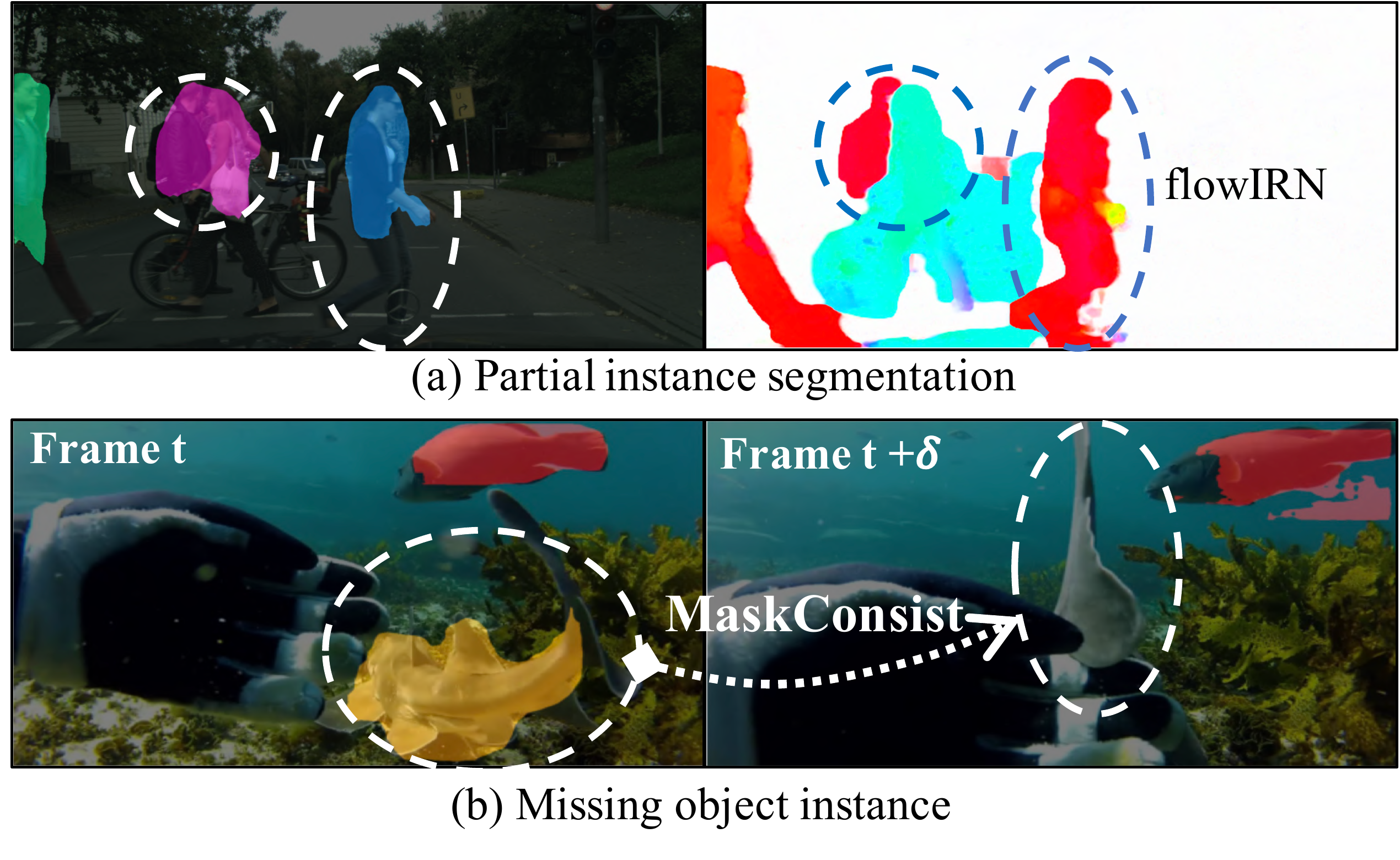}
\vspace{-0.03\linewidth}
\caption{Two types of error for IRN~\cite{ahn2019weakly} trained with still images: (a) partial segmentation and (b) missing instance. We observe optical flow is able to capture pixels of the same instance better (circles in (a)) and we propose flowIRN to model this information. In (b), a \textit{fish} is missed on one frame. We propose MaskConsist to leverage temporal consistency and transfer stable mask predictions to neighboring frames during training. }
\label{fig:pull_figure}
\vspace{-0.07\linewidth}
\end{figure}

Alternatively, weak labels like classification labels can be used to train instance segmentation models~\cite{zhou2018weakly,cholakkal2019object,zhu2019learning,ahn2019weakly,laradji2019masks,ge2019label,shen2019cyclic,arun2020weakly}. While weak labels are significantly cheaper to annotate, training weakly supervised models can be far more challenging. They typically suffer from two sources of error: (a) \textit{partial instance segmentation} and (b) \textit{missing object instances}, as shown in Fig.~\ref{fig:pull_figure}. Weakly supervised methods often identify only the most discriminative object regions that help predict the class label. This results in \textit{partial segmentation of objects}, as shown in Fig.~\ref{fig:pull_figure}(a). For instance, the recent work on weakly supervised instance segmentation IRN~\cite{ahn2019weakly} relies on class activation maps (CAMs)~\cite{zhou2016learning}, which suffer from this issue as also observed in other works~\cite{kolesnikov2016seed,wei2017object,zhang2018adversarial}. Further, CAMs do not differentiate between overlapping instances of the same class. It can also \textit{miss object instances} when multiple instances are present in an image, as shown in Fig.~\ref{fig:pull_figure}(b). In particular, an instance could be segmented in one image but not in another image where it is occluded or its pose alters.

Interestingly, these issues are 
less severe in videos, where object motion provides an additional signal for instance segmentation. As shown in Fig.~\ref{fig:pull_figure}, optical flow in a video is tightly coupled with instance segmentation masks. 
This is unsurprising since pixels belonging to the same (rigid) object move together and have similar flow vectors. 
We incorporate such video signals to train weakly supervised instance segmentation models, in contrast to existing methods~\cite{ahn2019weakly,laradji2019masks,shen2019cyclic,arun2020weakly} only targeted at images.

Typical weakly supervised approaches involve two steps: (a) generating pseudo-labels, comprising noisy instance segmentation masks consistent with the weak class labels, and (b) training a supervised model like Mask R-CNN based on these pseudo-labels. We leverage video information in both stages. In the first step, we modify IRN to assign similar labels to pixels with similar motion. This helps in addressing the problem of partial segmentation. We refer to the modified IRN as \textit{flowIRN}. In the second step, we introduce a new module called \textit{MaskConsist}, which counters the problem of missing instances by leveraging temporal consistency between objects across consecutive frames. It matches prediction between neighboring frames and transfers the stable predictions to obtain additional pseudo-labels missed by \textit{flowIRN} during training. This is a generic module that can be used in combination with any weakly supervised segmentation methods as we show in our experiments.

To the best of our knowledge, we are the first work to utilize temporal consistency between frames to train a weakly supervised instance segmentation model for videos. We show that this leads to more than $5\%$ and $3\%$ improvement in average precision compared to image-centric methods, like IRN, on video frames from two challenging video datasets: Youtube-VIS (YTVIS)~\cite{yang2019video} and Cityscapes~\cite{Cordts2016Cityscapes}, respectively. We also observe similar gains on the recently introduced video instance segmentation task~\cite{yang2019video} in YTVIS.
\vspace{-0.08\linewidth}
\section{Related Work}
\vspace{-0.02\linewidth}
Different types of weak supervision have been used in the past for semantic segmentation: bounding boxes~\cite{dai2015boxsup,papandreou2015weakly,khoreva2017simple,song2019box}, scribbles~\cite{lin2016scribblesup,vernaza2017learning,tang2018normalized}, and image-level class labels~\cite{kolesnikov2016seed,jin2017webly,hou2018self,wei2018revisiting,huang2018weakly,lee2019ficklenet,shimoda2019self,sun2020mining}. Similarly, for instance segmentation, image-level~\cite{zhou2018weakly,cholakkal2019object,zhu2019learning,ahn2019weakly,laradji2019masks,ge2019label,shen2019cyclic,arun2020weakly} and bounding box supervision~\cite{khoreva2017simple,hsu2019bbtp} have been explored. In this work, we focus on only using class labels for weakly supervised instance segmentation. 

\textbf{Weakly supervised semantic segmentation:}
Most weakly supervised semantic segmentation approaches rely on class attention maps (CAMs)~\cite{zhou2016learning} to provide noisy pseudo-labels as supervision~\cite{kolesnikov2016seed,huang2018weakly,shimoda2019self,sun2020mining}. Sun \textit{et al.}~\cite{sun2020mining} used co-attention maps generated from image pairs to train the semantic segmentation network. Another line of work leverages motion and temporal consistency in videos~\cite{tsai2016semantic,tokmakov2016learning,saleh2017bringing,hong2017weakly,lee2019frame,wang2020deep} to learn more robust representation. For instance, frame-to-frame (F2F)~\cite{lee2019frame} used optical flow to warp CAMs from neighboring frames and aggregated the warped CAMs to obtain more robust pseudo-labels.

\textbf{Weakly supervised instance segmentation:}
For training instance segmentation models with bounding box supervision, Hsu \textit{et al.}~\cite{hsu2019bbtp} proposed a bounding box tightness
constraint and multiple instance learning (MIL) based objective. Another line of work that only uses class labels extracts semantic responses from CAMs or other attention maps and then combines them with object proposals~\cite{rother2004grabcut,pont2016multiscale} to generate instance segmentation masks ~\cite{zhou2018weakly,cholakkal2019object,zhu2019learning,laradji2019masks,shen2019cyclic}. However, these methods' performance heavily depends on the quality of proposals used, which are mostly pre-trained on other datasets.
Shen \textit{et al.}~\cite{shen2019cyclic} extracted attention maps from a detection network and then jointly learn the detection and segmentation networks in a cyclic manner. Arun \textit{et al.}~\cite{arun2020weakly} proposed a conditional network to model the noise in weak supervision and combined it with object proposals to generate instance masks.
The first end-to-end network (IRN)~\cite{ahn2019weakly} was proposed by Ahn \textit{et al.} to directly predict instance offset and semantic boundary which were combined with CAMs to generate instance mask predictions. Our method adapts \cite{ahn2019weakly} for the first step of training and combines it with a novel MaskConsist module. However, other weakly supervised can also be integrated into our framework if code is available. 

\textbf{Segmentation in videos:} A series of approaches have emerged for segmentation in videos~\cite{Bertasius_2020_CVPR,hu2019learning,athar2020stem,mohamed2020instancemotseg,liang2020polytransform}. Some works proposed to leverage the video consistency~\cite{vondrick2018tracking,wang2019learning,lu2020learning,luiten2020unovost}. Recently, Yang \textit{et al.}~\cite{yang2019video} extended the traditional instance segmentation task from images to videos and proposed Video Instance Segmentation task (VIS). VIS aims to simultaneously segment and track all object instances in the video. Every pixel is labeled with a class label and an instance track-ID. MaskTrack~\cite{yang2019video} added a tracking-head to Mask R-CNN~\cite{he2017mask} to build a new model for this task. Bertasius \textit{et al.}~\cite{Bertasius_2020_CVPR} improved MaskTrack by proposing a mask propagation head. This head propagated instance features across frames in a clip to get more stable predictions. To the best of our knowledge, there has been no work that has explored weakly supervised learning for the video instance segmentation task. We evaluate our method on this task by combining it with a simple tracking approach.  

%-------------------------------------------------------------------------
\vspace{-0.02\linewidth}
\section{Approach}
\vspace{-0.02\linewidth}
\begin{figure*}[t]
\centering
\includegraphics[width=0.94\linewidth]{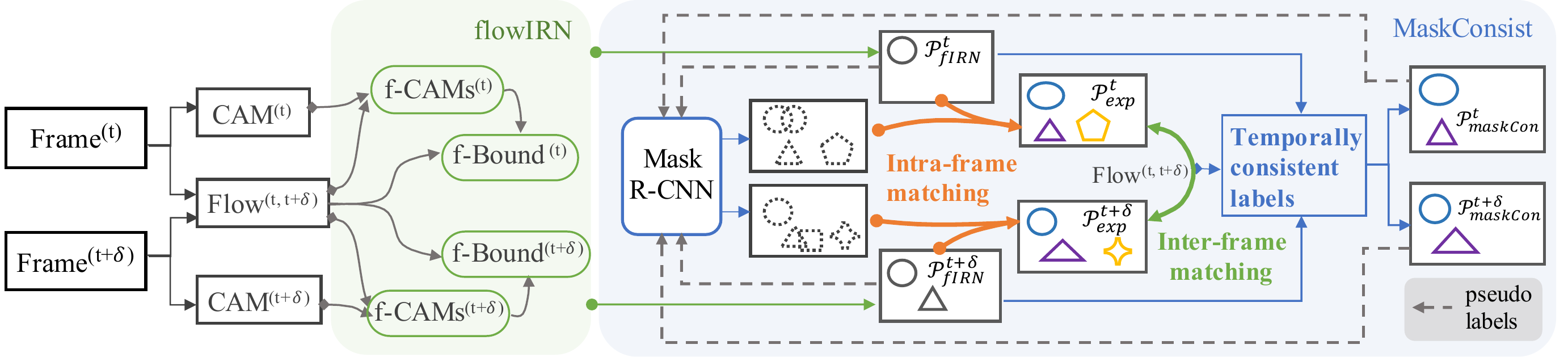}
\caption{Our pipeline mainly consists of two modules: flowIRN and MaskConsist. FlowIRN adapts IRN~\cite{ahn2019weakly} by incorporating optical flow to modify CAMs (f-CAMs), as well as introducing a new loss function: flow-boundary loss (f-Bound loss). MaskConsist matches the predictions from two successive frames and transfers high-quality predictions from one frame as pseudo-labels to another. It has three components: intra-frame matching, inter-frame matching and temporally consistent labels, shown in orange, green and blue, respectively. First, flowIRN is trained with frame-level class labels. Next, MaskConsist is trained with the pseudo-labels generated by flowIRN.}
\label{fig:overview}
\vspace{-0.03\linewidth}
\end{figure*}

We first introduce preliminaries of inter-pixel relation network (IRN)~\cite{ahn2019weakly} and extend it to incorporate video information, resulting in flowIRN. Next, we introduce MaskConsist which enforces temporal consistency in predictions across successive frames. Our framework has a 2-stage training process: (1) train flowIRN and (2) use masks generated by flowIRN on the training frames as supervision to train the MaskConsist model, as shown in Fig.~\ref{fig:overview}.

\vspace{-0.02\linewidth}
\subsection{Preliminaries of IRN}
\vspace{-0.02\linewidth}
IRN~\cite{ahn2019weakly} extracts inter-pixel relations from Class Attention Maps (CAMs) and uses it to infer instance locations and class boundaries. For a given image, CAMs provide pixel-level scores for each class that are then converted to class labels. Every pixel is assigned the label corresponding to the highest class activation score at the pixel, if this score is above a foreground threshold. Otherwise, it is assigned the background label.

IRN is a network with two branches that predict (a) a per-pixel \textit{displacement vector} pointing towards the center of the instance containing the pixel and (b) a per-pixel \textit{boundary likelihood} indicating if a pixel lies on the boundary of an object or not. Since the model is weakly supervised, neither displacement nor boundary labels are available during training. Instead, IRN introduces losses that enforce constraints on displacement and boundary predictions based on the foreground/background labels inferred from CAMs.

During inference, a two-step procedure is used to obtain instance segmentation masks. First, all pixels with displacement vectors pointing towards the same centroid are grouped together to obtain per-pixel instance labels. However, these predictions tend to be noisy. In the second step, the predictions are refined using a pairwise affinity term $\alpha$. For two pixels $i$ and $j$,
\begin{align}
\vspace{-0.01\linewidth}
    \small
    \alpha_{i,j}=1-\max_{k\in\Pi_{i,j}}\mathcal{B}(k), \label{eq:affinity}
\vspace{-0.02\linewidth}
\end{align}
where $\mathcal{B}(k)$ is the boundary likelihood predicted by IRN for pixel $k$, and $\Pi_{i,j}$ is the set of pixels lying on the line connecting $i$ and $j$. If two pixels are separated by an object boundary, at least one pixel on the line connecting them should belong to this boundary. This results in low affinity between the two pixels. Conversely, the affinity would be high for pixels which are part of the same instance. In IRN, the affinity term is used to define the transition probability for a random walk algorithm that smooths the final per-pixel instance and class label assignments.

\vspace{-0.02\linewidth}
\subsection{FlowIRN Module}
\label{sec:flowirn}
\vspace{-0.02\linewidth}
We introduce flowIRN which improves IRN by incorporating optical flow information in two components, flow-amplified CAMs and flow-boundary loss.

\textbf{Flow-Amplified CAMs:} We observed that CAMs identify only the discriminative regions of an object (like the face of an animal) but often miss other regions corresponding to the object. This has been noted in previous works as well~\cite{kolesnikov2016seed,wei2017object,zhang2018adversarial}. 
% Due to scale variations and motion blur, CAMs may miss some instances when there are multiple objects in the scene. 
Since the objects of interest in a video are usually moving foreground objects, we address this issue by first amplifying CAMs in regions where large motion is observed. More specifically, given the estimated optical flow $\mathcal{F}\in\mathbb{R}^{H\times W\times 2}$ for the current frame, we replace CAMs used in IRN with:
\begin{align}
\small
    \text{f-CAM}_c(x)=\text{CAM}_c(x)\times A^{\mathbb{I}(||\mathcal{F}(x)||_2>T)}, \label{eq:fCAMs}
\vspace{-0.04\linewidth}
\end{align}where $A$ is an amplification coefficient and $T$ is a flow magnitude threshold. This operation is applied to CAMs of all classes equally, preserving the relative ordering of class scores. Class labels obtained from CAMs are not flipped; only foreground and background assignments are affected.

\textbf{Flow-boundary loss:} In IRN, boundary prediction is supervised by the pseudo segmentation labels from CAMs, which does not distinguish instances of the same class, particularly overlapping instances. However, in videos, optical flow could disambiguate such instances, since pixels of the same rigid object move together and have consistent motion. Hence, we use spatial gradient of optical flow to identify if two pixels are from the same object instance or not. Points from the same object can be from different depths relative to the camera, and might not have the same optical flow. In practice, we observed that the gradient is more robust to this depth change. We explain this in detail in the appendix. We use the affinity term from Eq.~\ref{eq:affinity} to define a new flow-boundary loss:
\begin{align}
\small
    \mathcal{L}^{\mathcal{B}}_\mathcal{F} = \sum_{j\in\mathcal{N}_i} ||\mathcal{F}'(i)-\mathcal{F}'(j)||\alpha_{i,j}+\lambda|1-\alpha_{i,j}|, \label{eq:fBoundary}
\vspace{-0.04\linewidth}
\end{align}where $\mathcal{F'}(i)$ is a two-dimensional vector denoting the gradient of the optical flow at a pixel $i$ with respect to its spatial co-ordinates $(x_i,y_i)$ respective. $\mathcal{N}_i$ is a small pixel neighborhood around $i$ as defined in IRN, and $\lambda$ is the regularization parameter. The first term implies that pixels with similar flow-gradients could have high-affinity (belonging to the same instance), while pixels with different flow-gradients should have low-affinity (belonging to different instances). The second term is used to regularize the loss and prevent the trivial solution of $\alpha$ being $0$ constantly. We train flowIRN with the above loss and the original losses in IRN.
\vspace{-0.06\linewidth}
\subsection{MaskConsist}
\label{sec:maskconsist}
\vspace{-0.04\linewidth}
\begin{figure}[h]
\centering
\includegraphics[width=0.96\linewidth]{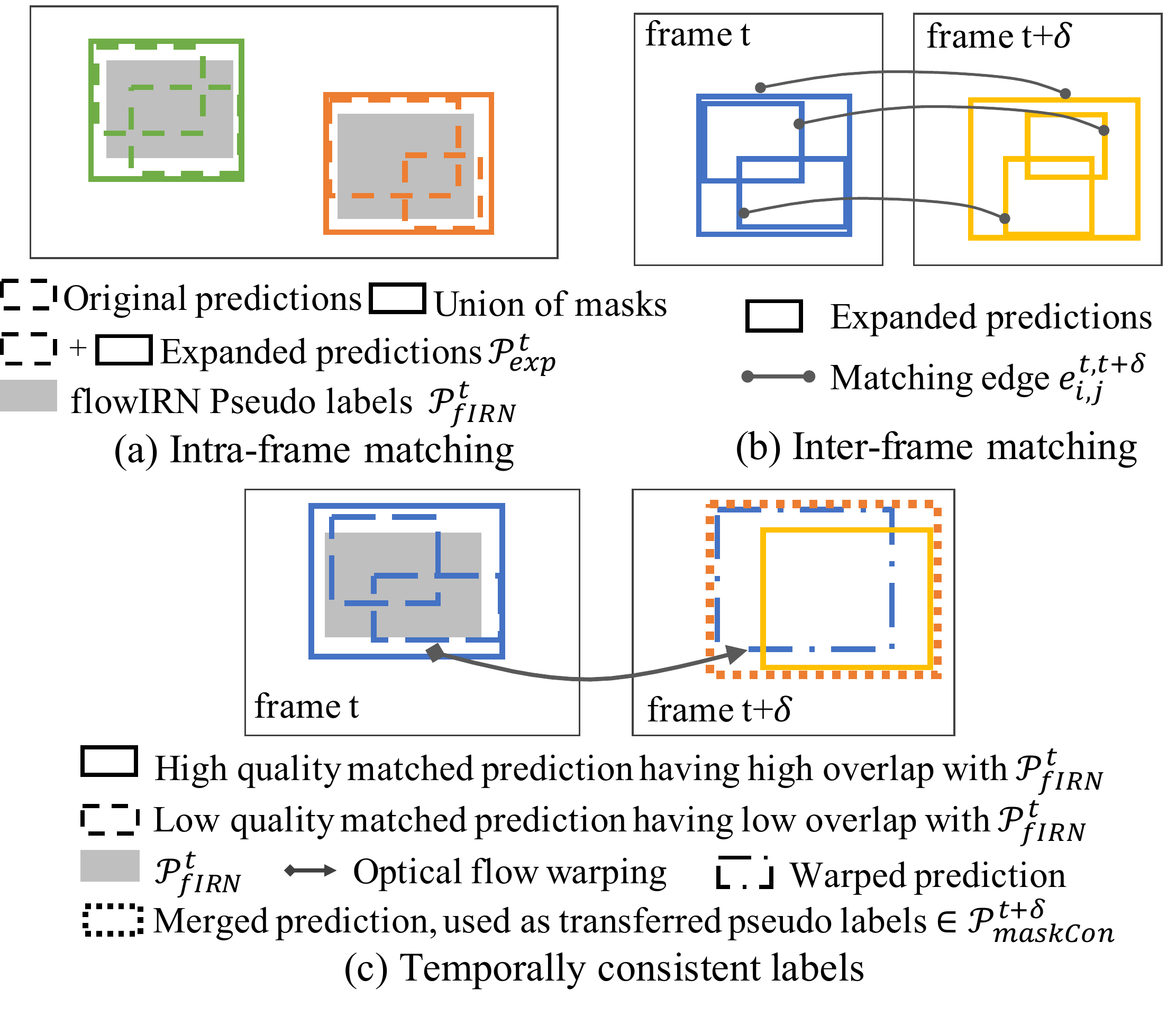}
\caption{Three steps of MaskConsist module. (a) Intra-frame matching expands the original predictions of current Mask R-CNN by merging highly overlapping predictions. (b) Inter-frame matching identifies one-to-one matching between predictions across frames. (c) Temporally consistent labels transfer matched predictions from one frame to another after warping with optical flow.} 
\label{fig:mconsist}
\vspace{-0.03\linewidth}
\end{figure}

The instance-level masks generated by flowIRN can now be used as pseudo-labels to train a fully supervised instance segmentation model, like Mask R-CNN. In practice, this yields better performance on the validation set than the original flowIRN model. However, the pseudo-labels generated by flowIRN can miss object instances on some frames if both CAMs and optical-flow couldn't identify them.

MaskConsist solves this by transferring ``high-quality" mask predictions from a neighboring frame as new pseudo-labels to the current frame while training a Mask R-CNN. At each training iteration, we train the network with a pair of neighboring frames $t$ and $t+\delta$ in the video. In addition to the pseudo-labels from flowIRN, a subset of predictions by the current Mask R-CNN on $t+\delta$ are used as additional pseudo-labels for $t$ and vice-versa. Predictions are transferred only if (a) they are temporally stable and (b) overlap with existing pseudo-labels form flowIRN. This avoids false-negatives by MaskConsist. MaskConsist contains three steps: intra-frame matching, inter-frame matching, and temporally consistent label assignment. These steps are explained next and visualized in Fig.~\ref{fig:mconsist}.

\textbf{Intra-frame matching:} At each training iteration, we first generate a large set of candidate masks that can be transferred to neighboring frames. The pseudo-labels from flowIRN might be incomplete, but we expect the predictions from the Mask R-CNN to become more robust as training proceeds. Hence, high-confidence predictions from the model at a given iteration can be used as the candidate set. However, we empirically observed that during early stages of training, masks predicted by Mask R-CNN can be fragmented. We overcome this by also considering the union of all mask predictions which have good overlap  with a flowIRN pseudo-label of the same class. The candidate set of predictions after this step includes the original predictions (in practice, we use top 100 predictions) for the frame, as well as the new predictions obtained by combining overlapping predictions, as shown in Fig.~\ref{fig:mconsist}(a). For a frame at time $t$, we refer to the original set of predictions from the model as $\mathcal{P}^t$, and this expanded set as $\mathcal{P}_{\text{exp}}^{t}$. Each prediction $p^t_i \in \mathcal{P}_{\text{exp}}^{t}$ corresponds to a triplet: mask, bounding box and the class with highest score for the box, denoted by $\left(m^t_i, b^t_i, c^t_i\right)$ respectively.

\textbf{Inter-frame matching:} Next, we wish to transfer some predictions from the current frame $t$ as pseudo-labels to the neighboring frame $t+\delta$ and vice-versa. We only transfer a prediction if it is stably predicted by the current model on both frames. To do this, we first create a bipartite graph between the two frames. The nodes from each frame correspond to the expanded prediction set $\mathcal{P}_{\text{exp}}^{t}$ and $\mathcal{P}_{\text{exp}}^{t + \delta}$ respectively as shown in Fig.~\ref{fig:mconsist}(b). The edge weight $e_{ij}^{t, t+\delta}$ between prediction $p^t_i$ and $p^{t+\delta}_j$ is defined as:
\begin{align*}
\small
    & e_{ij}^{t, t+\delta} = \mathbb{I}( c_{i}^{t}= c_{j}^{t+\delta})\cdot\text{IoU}(\mathrm{W}_{t\rightarrow t+\delta}(p_i^{t}), p_{j}^{t+\delta}),
    \vspace{-0.02\linewidth}
\end{align*}
where $\mathrm{W}_{t\rightarrow t+\delta}$ is a bi-linear warping function that warps the prediction from one frame to another based on the optical flow between them (explained in the appendix). The edge weight is non-zero only if the two predictions share the same class. The weight is high if the warped mask from frame $t$ has high overlap with the mask in $t+\delta$.

The correspondence between predictions of both frames is then obtained by solving the bipartite graph-matching problem with these edge weights, using the Hungarian algorithm. This results in a one-to-one matching between a subset of predictions from $\mathcal{P}_{\text{exp}}^{t}$ and $\mathcal{P}_{\text{exp}}^{t + \delta}$. We denote the matching result as $\mathcal{M}^{t, t + \delta} = \{(p_i^t, p_j^{t+\delta})\}$, containing pairs of matched predictions from both frames. This comprises pairs of predictions that are temporally stable.

\textbf{Temporally consistent labels:} We use the predictions from frame $t$ which are matched to some predictions in $t + \delta$ in the previous step to define new pseudo-labels for frame $t + \delta$ as shown in Fig.~\ref{fig:mconsist}(c). Since there can be a lot of spuriously matched predictions, we only transfer high-quality predictions that have some overlap with the original pseudo-labels in frame $t$. As the process presented in Alg.~\ref{algo:cons}, let $\mathcal{P}_{\text{fIRN}}^{t}$ be the original set of pseudo-labels obtained from flowIRN for frame $t$ and $\mathcal{M}^{t, t + \delta}$ be the matched prediction pairs between two frames. We transfer only those masks from $t$ which have an overlap greater than $0.5$ with any of the original masks in $\mathcal{P}_{\text{fIRN}}^{t}$ of the same class. Further, when transferring to $t + \delta$, we (a) warp the mask using optical flow and (b) merge it with the matched prediction in $t+\delta$ as shown in Fig.~\ref{fig:mconsist}(c) to ensure that the mask is not partially transferred. This new set of labels transferred from $t$ to $t + \delta$ are denoted by $\mathcal{P}_{\text{maskCon}}^{t + \delta}$. The steps are explained below. Here, $Merge(\cdot)$ simply takes the union of masks from two predictions to form a new prediction. 
\vspace{-0.03\linewidth}
\begin{algorithm}
\caption{Temporally consistent assignment}\label{algo:cons}
\small
    \KwIn{$\mathcal{M}^{t, t + \delta}, \;\; \mathcal{P}_{\text{fIRN}}^t$}
    \KwOut{$\mathcal{P}_{\text{maskCon}}^{t + \delta}$}
    $\mathcal{P}_{\text{maskCon}}^{t + \delta} \gets \{\}$ \\
    \For{$(p_i^t, p_j^{t+\delta}) \in \mathcal{M}^{t, t + \delta}$}{
        \For{$p_{\text{fIRN}}^t \in \mathcal{P}_{\text{fIRN}}^t$}{
            \uIf{$\text{IoU}(b_i^t, b_{\text{fIRN}}^t) > 0.5$, $c_i^t = c_{\text{fIRN}}^t$} {
            $p^{t + \delta}_m \gets \text{Merge}(\mathrm{W}_{t\rightarrow t+\delta}(p_i^t), p_j^{t + \delta})$ \\
            $\mathcal{P}_{\text{maskCon}}^{t + \delta} \gets \mathcal{P}_{\text{maskCon}}^{t + \delta} \cup \{p^{t + \delta}_m\}$ \\
            \textbf{break}
            }
        }
    }
    \Return $\mathcal{P}_{\text{maskCon}}^{t + \delta}$
\end{algorithm}
\vspace{-0.04\linewidth}

Simultaneously, new pseudo-labels $\mathcal{P}_{\text{maskCon}}^t$ are obtained for $t$ by transferring predictions from $t + \delta$ in a similar fashion. 
We combine them with the original pseudo-labels from flowIRN to obtain the final set of pseudo-labels $\mathcal{P}_{\text{maskCon}} \cup \mathcal{P}_{\text{fIRN}}$. We also note that while combining these two sets of labels, it is important to suppress smaller masks that are contained within the others. Concretely, we apply non-maximal suppression (NMS) based on an Intersection over Minimum (IoM) threshold. IoM is calculated between two masks as the intersection area over the area of the smaller mask. This avoids label redundancy and helps improve performance as we demonstrate later in ablation experiments. The merged pseudo-labels are used as supervision to train the Mask R-CNN model as shown in Fig.~\ref{fig:overview}, without altering the Mask R-CNN in any way.

Our overall MaskConsist approach does not require extra forward or backward pass, and only adds a small overhead to the original Mask R-CNN during training. During inference, the matching is unnecessary and MaskConsist works similar to Mask R-CNN.

%-------------------------------------------------------------------------
\vspace{-0.02\linewidth}
\section{Experiments}
\vspace{-0.02\linewidth}
Unless otherwise specified, models in this section are trained only with frame-level class labels and do not use bounding-box or segmentation labels. We evaluate our model on two tasks: frame-level instance segmentation and video-level instance segmentation. We report performances on two popular video datasets. 

\vspace{-0.02\linewidth}
\subsection{Datasets}
\vspace{-0.02\linewidth}
\textbf{Youtube-VIS} (YTVIS)~\cite{yang2019video} is a recently proposed benchmark for the task of video instance segmentation. It contains $2,238$ training, $302$ validation, and $343$ test videos collected from YouTube, containing $40$ categories. Every $5$th frame in the training split is annotated with instance segmentation mask.

As the annotation of validation and test splits are not released and only video-level instance segmentation performance is available on the evaluation server, we hold out a subset of videos from the original training split by randomly selecting $10$ videos from each category. This results in a train\_val split of $390$ videos (there are videos belonging to multiple object categories) to conduct frame-level and video-level instance segmentation evaluations.
% with $10627$ frames, 
The remaining $1,848$ videos 
% with $50,714$ frames 
are used as the train\_train split.

\textbf{Cityscapes}~\cite{Cordts2016Cityscapes} contains high-quality pixel-level annotations for $5,000$ frames collected in street scenes from $50$ different cities. $19$ object categories are annotated with semantic segmentation masks and $8$ of them are annotated with instance segmentation masks. The standard $2,975$ training frames and their neighboring $t-3$ and $t+3$ frames are used for training, and the $500$ frames in validation split are used for evaluation.

\vspace{-0.02\linewidth}
\subsection{Implementation details}
\label{sec:impl}
\vspace{-0.02\linewidth}
\textbf{Optical flow network:} We use the self-supervised DDFlow~\cite{liu2019ddflow} for optical flow extraction. The model is pre-trained on ``Flying Chairs'' dataset~\cite{DFIB15} and then fine-tuned on YTVIS or Cityscapes training videos in an unsupervised way. The total training time is $120$ hours on four P100 GPUs and the average inference time per frame is 280ms.

\textbf{flowIRN:} To get flow-amplified CAMs, we set the amplification co-efficient $A=2$ and threshold $T=\text{Percentile}_{0.8}(||\mathcal{F}(x)||_2)$ for YTVIS, and $A=5$ and $T=\text{Percentile}_{0.5}(||\mathcal{F}(x)||_2)$ for Cityscapes. The optical flow is extracted between two consecutive frames (frame $t$ and $t+1$). The regularization weight $\lambda$ is set to $2$. We train the network for $6$ epochs. Other training and inference hyper-parameters are set the same as in~\cite{ahn2019weakly}. Empirically, we observe that IRN (and flowIRN) is limited by lack of good CAMs when trained only on Cityscapes data. Hence, for experiments on Cityscapes, we train the first-stage of all weakly supervised models (before the Mask R-CNN/MaskConsist stage) first on PASCAL VOC 2012~\cite{everingham2010pascal} training-split and then fine-tune on Cityscapes.

\textbf{MaskConsist:} We use ResNet-50 as the backbone, initialized with ImageNet pre-trained weights. For both datasets, the bounding-box IoU threshold is set at $0.5$ for intra-frame matching, and IoM-NMS threshold at $0.5$ for label combining. The model is trained for $90K$ iterations for YTVIS, and $75K$ iterations for Cityscapes, with base learning rate $lr=0.002$. SGD optimizer is used with step schedule $\gamma=0.1$, decay at $75\%$ and $88\%$ of total steps. The temporal consistency is calculated between frame $t$ and $t+5$ ($\delta=5$) for YTVIS, frame $t$ and $t+3$ ($\delta=3$) for Cityscapes. Inference on one frame (short side $480$px) takes 210ms. Nvidia Tesla P100 GPU is used in training and test. All hyper-parameters for flowIRN and MaskConsist are selected based on the performance on a small held-out validation split of the corresponding training set.

\begin{table}[]
\centering
\small
\setlength{\tabcolsep}{1mm}
\begin{tabular}
{|@{\hspace{\tabcolsep}}L{28mm}|C{16mm}|C{18mm}|C{12mm}|C{22mm}|}
\hline
Methods       & Video Info & Supervision & $AP_{50}$\\ \hline
Mask R-CNN~\cite{he2017mask}     &    \xmark   &  Mask   & $78.24$ \\ 

WSIS-BBTP~\cite{hsu2019bbtp}     &         \xmark   & Bbox        &   $46.80$                \\\hline
WISE~\cite{laradji2019masks}          &         \xmark & Class       & $24.54$             \\
F2F~\cite{lee2019frame}+MCG~\cite{pont2016multiscale}    & \cmark & Class       & $26.31$             \\
IRN~\cite{ahn2019weakly}           &          \xmark & Class       & $29.64$       \\ 
IRN~\cite{ahn2019weakly}+F2F\cite{lee2019frame}      & \cmark                     & Class       &       $30.27$            \\ \hline

Ours          & \cmark & Class       & $34.66$      \\ 
Ours (self-training)          & \cmark                    & Class       & $\textbf{36.00}$      \\ \hline
\end{tabular}
\caption{Frame-level instance segmentation performance ($AP_{50}$) on YTVIS train\_val split.}
\label{tab:ytvis_fis}
\vspace{-0.07\linewidth}
\end{table}

\textbf{Experiment setup:} 
On YTVIS, all methods are trained using the training frames (every $5$th frame) in train\_train split.
On Cityscapes, all methods are trained with training frames (frame $t$) and their two neighboring frames ($t-3$ and $t+3$).
Unless otherwise specified, our model is trained in two-steps: first train flowIRN on training frames, then use the pseudo-labels generated by the flowIRN on the training frames to train MaskConsist. For fair comparison, all baseline methods are also trained in two steps: first train the weakly supervised model (e.g.,~IRN) with frame-level class labels, then use pseudo-labels obtained to train a Mask R-CNN model. This is common practice in weakly supervised segmentation works~\cite{ahn2019weakly,laradji2019masks}, and improves $AP_{50}$ of all models by at least $2\%$ in our experiments. The same hyper-parameters reported in the original work or published code are retained for all baselines. 

We also observe that a three-step training process, where the masks generated by our MaskConsist model are used to train another MaskConsist model, further improves performance. We refer to this as \textit{ours self-training}. 
Note that unlike other baselines, this involves an additional round of training. 
On other baseline methods, we also attempted self-training: another round of training using pseudo-labels from the trained Mask R-CNN. However, this either degraded or did not improve performance on the validation set.

During frame-level inference, the trained MaskConsist or Mask R-CNN (for other baselines) is applied on each frame with score threshold of $0.05$ and NMS threshold of $0.5$ to obtain prediction masks. For video-level evaluation, we apply an unsupervised tracking method~\cite{luiten2020unovost} on per-frame instance mask predictions to obtain instance mask tracks, with the same hyper-parameters as the original work.
We will release our code after paper acceptance.

\begin{table}[]
\centering
\small
\setlength{\tabcolsep}{1mm}
\begin{tabular}{|l|c|c|c|}
\hline
Methods     & Supervision     & Instance seg & Semantic seg \\ \hline
Mask R-CNN~\cite{he2017mask}  & Mask   & $38.73$   & $79.23$ \\ \hline
WISE~\cite{laradji2019masks}  & Class       &         $10.51$              &         $35.82$              \\
F2F~\cite{lee2019frame}+MCG~\cite{pont2016multiscale}  & Class & $10.73$   &  $33.26$ \\
IRN~\cite{ahn2019weakly}   & Class   &          $12.33$             &      $33.48$                 \\
IRN~\cite{ahn2019weakly}+F2F\cite{lee2019frame} & Class & $12.53$   &  $34.17$ \\ \hline
Ours        & Class     &          $16.05$             &      $39.88$                 \\ 
Ours (self-training) & Class     &          $\textbf{16.82}$             &      $\textbf{41.31}$  \\ \hline 

\end{tabular}
\caption{Frame-level instance segmentation ($AP_{50}$) and semantic segmentation (\textit{IoU}) on Cityscapes validation split.}
\label{tab:city}
\vspace{-0.06\linewidth}
\end{table}

\vspace{-0.02\linewidth}
\subsection{Frame instance segmentation}
\vspace{-0.02\linewidth}
First, we compare frame-level performance with existing instance segmentation models on YTVIS and Cityscapes.

\textbf{Evaluation metrics:} On both YTVIS and Cityscapes, the average precision with mask intersection over union (\textit{IoU}) threshold at $0.5$ ($AP_{50}$) is used as the metric for instance segmentation. Cityscapes is a popular benchmark for semantic segmentation and we also report the semantic segmentation performance using standard \textit{IoU} metric.

\begin{table*}[]
\centering
\small
\setlength{\tabcolsep}{1mm}
\begin{tabular}{|C{24mm}|L{23mm}|C{8mm}|C{8mm}|C{8mm}|C{8mm}|C{8mm}|C{8mm}|C{8mm}|C{8mm}|C{8mm}|C{8mm}|}
\hline
\multicolumn{2}{|c|}{\multirow{2}{*}{Methods}}                                     & \multicolumn{5}{c|}{Train\_Val Split} & \multicolumn{5}{c|}{Validation Split} \\ \cline{3-12} 
\multicolumn{2}{|c|}{}                                                             & $mAP$ & $AP_{50}$ & $AP_{75}$ & $AR_{1}$ & $AR_{10}$      & $mAP$  & $AP_{50}$ & $AP_{75}$ & $AR_{1}$ & $AR_{10}$    \\ \hline
\multirow{3}{*}{\begin{tabular}{@{}c@{}}Fully supervised \\ learning methods\end{tabular}}  & IoUTracker+~\cite{yang2019video}                      & -          & -   & -   & -   & -    & $23.6$     & $39.2$    & $25.5$ & $26.2$ & $30.9$    \\ 
 & DeepSORT~\cite{wojke2017simple}    & -    & -   & -          & -          & -         & $26.1$     & $42.9$  & $26.1$ & $27.8$   & $31.3$    \\ 
 & MaskTrack~\cite{yang2019video}   & -    & -     & -          & -          & -         & $30.3$     & $51.1$   & $32.6$ & $31.0$ & $35.5$    \\ \hline
\multirow{3}{*}{\begin{tabular}{@{}c@{}}Weakly supervised \\ learning methods\end{tabular}} &  WISE~\cite{laradji2019masks}                  &    $8.7$        &     $22.1$       & $5.5$ & $9.8$ &   $10.7$      &      $6.3$    &     $17.5$    &    $3.5$ &  $7.1$ & $7.8$    \\ 
& IRN~\cite{ahn2019weakly}                  &   $10.8$         &      $26.4$    &  $7.7$ & $12.6$ &     $14.4$      &    $7.3$      &      $18.0$   & $3.0$  & $9.0$   &  $10.7$   \\ \cline{2-12} 
& Ours                 & $14.1$       & $34.4$    &  $9.4$ & $16.0$  & $17.9$      & $10.5$     & $27.2$   & $6.2$ & $12.3$ & $13.6$    \\ \hline
\end{tabular}
\caption{Video instance segmentation results on Youtube-VIS dataset.}
\label{tab:vis}
\vspace{-0.02\linewidth}
\end{table*}

\begin{figure*}[t]
\centering
\includegraphics[width=0.88\linewidth]{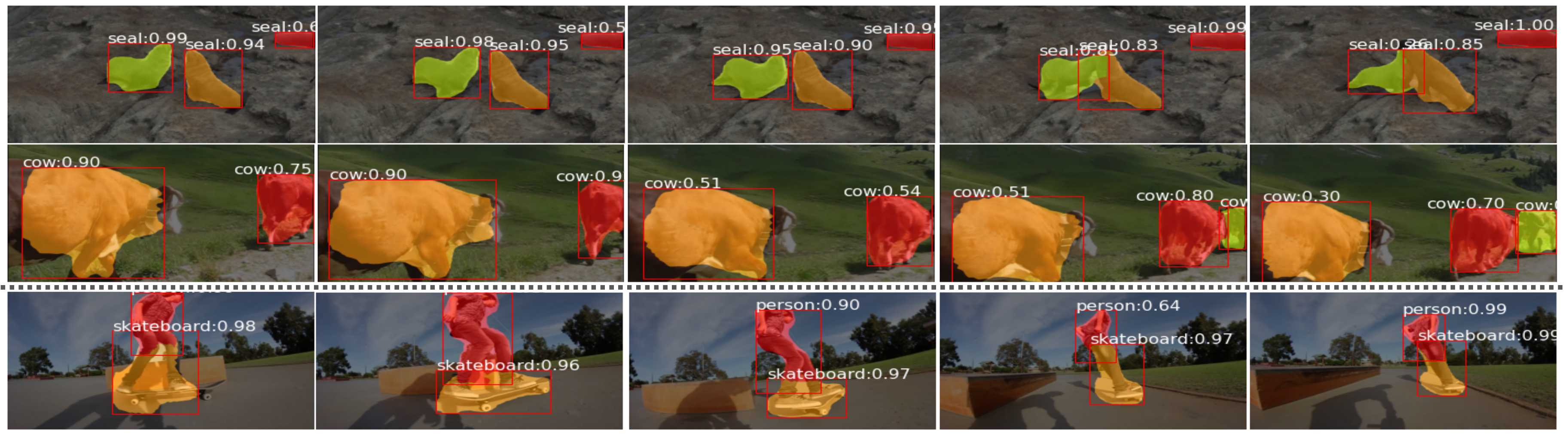}
\vspace{-0.01\linewidth}
\caption{Example Video instance segmentation results from our method on Youtube-VIS dataset.}
\label{fig:vis}
\vspace{-0.03\linewidth}
\end{figure*}

\textbf{Baselines:} To the best of our knowledge, there is no existing weakly supervised instance segmentation model designed for videos. Existing works are designed for still images and report results on standard image benchmarks like~\cite{everingham2010pascal}. To compare with these models on video data, we train them (where code is available) with independent video frames of YTVIS or Cityscapes. We also extend existing weakly supervised ``video" semantic segmentation models to perform instance segmentation. For upper-bound comparisons, we report results from Mask R-CNN~\cite{he2017mask} trained with ground truth masks, and WSIS-BBTP~\cite{hsu2019bbtp} trained with bounding box annotations. We list the baselines below and more details can be found in the appendix:
\vspace{-0.04\linewidth}
\begin{itemize}[leftmargin=*]
\setlength\itemsep{-0.4em}
    \item \textit{WISE~\cite{laradji2019masks}}: train on independent frame with class label.
    \item \textit{IRN~\cite{ahn2019weakly}}: train on independent frame with class label.
    \item \textit{F2F~\cite{lee2019frame} + MCG~\cite{pont2016multiscale}}: use videos with class labels to train F2F to obtain semantic segmentation and combine MCG proposals to obtain instance-level masks as in~\cite{zhou2018weakly}.
    \item \textit{F2F~\cite{lee2019frame} + IRN~\cite{ahn2019weakly}}: use optical flow to aggregate CAMs as in F2F to train IRN.
\end{itemize}
\vspace{-0.04\linewidth}

\textbf{Results:} Results on YTVIS are shown in Tab.~\ref{tab:ytvis_fis}.
All methods use two-step training as stated in the experiment setup.
WISE and F2F+MCG both use processed CAMs as weak labels and combine results with object proposals (MCG) to distinguish instances. 
Comparing WISE and F2F+MCG, F2F uses video information that boosts its performance by around $1.8\%$. IRN+F2F is the closest comparison to our approach, since it is also built on top of IRN and uses video information. Our model outperforms IRN+F2F by more than $4\%$, and can also benefit from an additional round of self-training (Ours self-training). 
However, we do not observe any gains when training the Mask R-CNN for another round for other methods.

In Tab.~\ref{tab:city}, we report frame-level instance segmentation and semantic segmentation results on Cityscapes. For instance segmentation, our method outperforms WISE and IRN by more than $3.7\%$ under $AP_{50}$. We convert the instance segmentation results to semantic segmentation by merging instance masks of the same class and assigning labels based on scores. On semantic segmentation, our method still outperforms IRN by a large margin. 

\vspace{-0.02\linewidth}
\subsection{Video instance segmentation}
\vspace{-0.02\linewidth}
Given per-frame instance segmentation predictions, we apply the Forest Path Cutting algorithm~\cite{luiten2020unovost} to obtain a mask-track for each instance and report VIS results.

\textbf{Evaluation metric:} We use the same metrics as \cite{yang2019video}: mean average precision for IoU between $[0.5, 0.9]$ ($mAP$), average precision with IoU threshold at $0.5$ / $0.75$ ($AP_{50}$/ $AP_{75}$), and average recall for top $1$ / $10$ ($AR_{1}$ / $AR_{10}$). As each instance in a video contains a sequence of masks, the computation of IoU uses the sum of intersections over the sum of unions across all frames in a video. The evaluation is carried out on YTVIS train\_val split using YTVIS code (\url{https://github.com/youtubevos})
, and also on YTVIS validation split using the official YTVIS server.
 
\textbf{Baselines:} Since there is no existing work on weakly supervised video instance segmentation, we construct our own baselines by combining the tracking algorithm in~\cite{luiten2020unovost} with two weakly supervised instance segmentation baselines: WISE~\cite{laradji2019masks} and IRN~\cite{ahn2019weakly}. We also present published results from fully supervised methods~\cite{yang2019video,wojke2017simple} for reference.

As presented in Tab.~\ref{tab:vis}, our model outperforms IRN and WISE by a large margin. On the $AP_{50}$ metric, there is a boost of more than $8\%$ on both train\_val and validation splits. We also observe that the performance gap between WISE and IRN decreases compared with frame-level results in Tab.~\ref{tab:ytvis_fis}, implying temporal consistency is important to realize gains in video instance segmentation. Note that the fully supervised methods are first trained on MS-COCO~\cite{lin2014microsoft} and then fine-tuned on YTVIS training split, while ours is only trained on YTVIS data. Qualitative VIS results from our method are shown in Fig.~\ref{fig:vis}. Our method generates temporally stable instance predictions and is able to capture different overlapping instances. One failure case is shown in the bottom row. As \textit{skateboard} and \textit{person} always appear and move together in YTVIS, our assumption on different instances having different motion is not valid. Thus, these two instances are not well distinguished.

\vspace{-0.02\linewidth}
\subsection{Effect of modeling temporal information}
\label{sec:diag}
\vspace{-0.02\linewidth}
Our framework explicitly models temporal information in both flowIRN and MaskConsist modules. We explore the effectiveness of each module in this section.

\begin{table}[t]
\centering
\small
\begin{tabular}{|L{22mm}|C{15mm}|C{15mm}|}
\hline
                  & YTVIS & Cityscapes \\ \hline
IRN~\cite{ahn2019weakly}    & $25.42$                    & $8.46$       \\ 
IRN+f-Bound    & $26.60$                      & $9.51$       \\
IRN+f-CAMs         & $27.47$                    & $10.55$      \\ \hline
flowIRN           & $\textbf{28.45}$                 & $\textbf{10.75}$      \\ \hline
\end{tabular}
\caption{Ablation study of flowIRN components. Results are reported on training data to evaluate pseudo-label quality. No second-step Mask R-CNN or MaskConsist training is applied here.}
\label{tab:flowIRN}
\vspace{-0.07\linewidth}
\end{table}

\textbf{Ablation study of flowIRN:} In Tab.~\ref{tab:flowIRN}, we present the instance segmentation results ($AP_{50}$) of different flowIRN variants. All models are directly tested on the training data to evaluate pseudo-label quality and no second-step training is used in this experiment. 
Compared to original IRN\cite{ahn2019weakly}, both flow-amplified CAMs (f-CAMs) and flow-boundary loss (f-Bound) incorporate optical flow information and improve IRN performance. Combining the two leads to our design of flowIRN, which improves by $3.03\%$ on YTVIS and $2.29\%$ on Cityscapes compared to IRN.

\begin{figure}[t]
\vspace{-0.01\linewidth}
\centering
\includegraphics[width=0.9\linewidth]{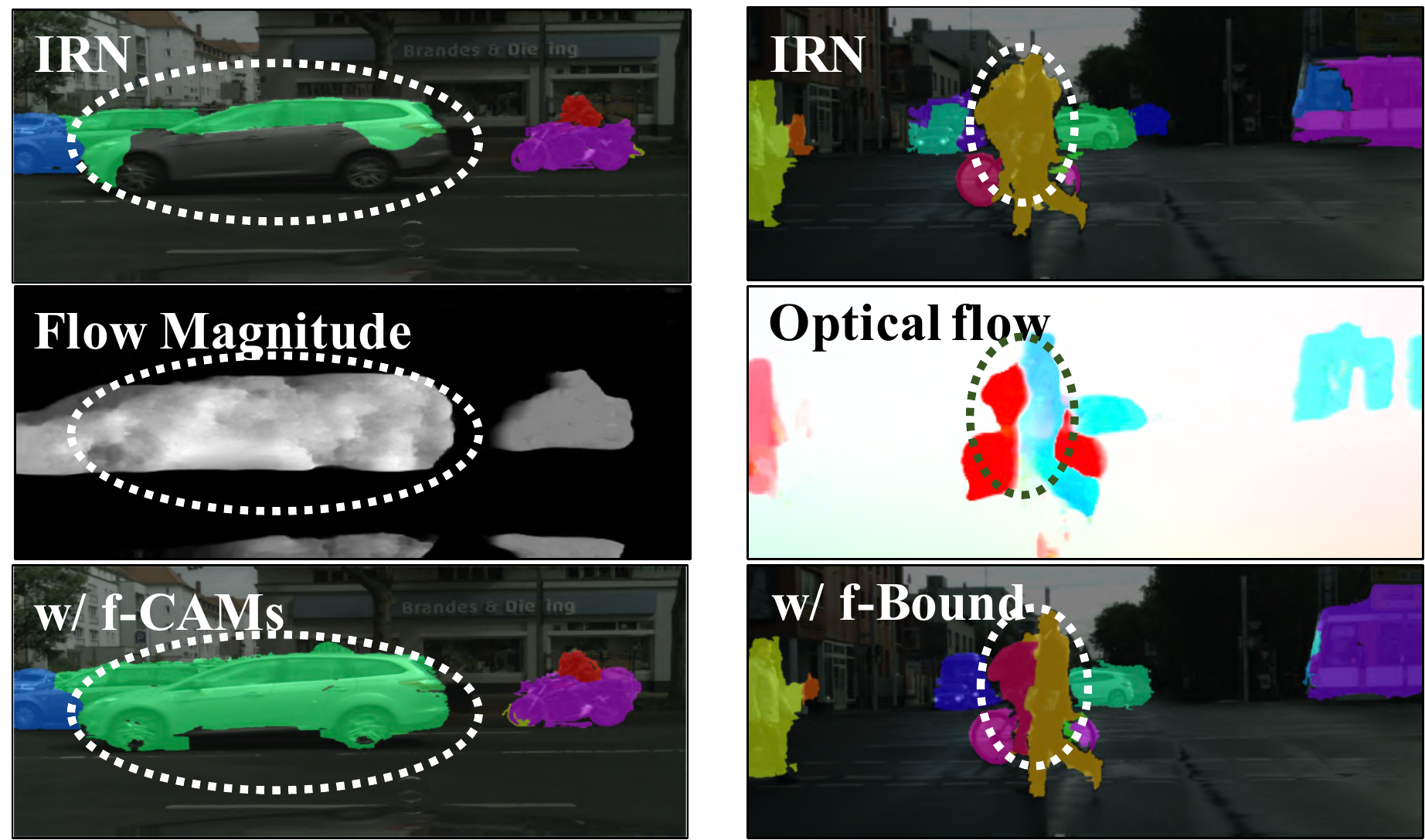}
\vspace{-0.01\linewidth}
\caption{Improvement introduced by f-CAMs and f-Bound. Top: output of IRN. Middle: optical flow extracted for the input frame. Bottom: output after incorporating f-CAMs or f-Bound.}
\label{fig:flowIRN}
\vspace{-0.04\linewidth}
\end{figure}

In Fig.~\ref{fig:flowIRN}, we show two qualitative examples of incorporating f-CAMs and f-Bound. In the left example, the car (in the circle) moves fast and is partially missed by IRN. After applying f-CAMs, the whole object is well captured in the segmentation mask. In the second example (right column), IRN fails to separate two overlapping persons while the boundary is recognizable in optical flow. After applying f-Bound loss, two instances are correctly predicted.

\textbf{Ablation study of MaskConsist:} In Tab.~\ref{tab:maskconsist}, we explore the contribution of different components of MaskConsist by disabling one of the three components each time. We observe that inter-frame matching plays the most important role in MaskConsist. It enables the model to incorporate temporal consistency during training and achieves the largest performance boost. IoM-NMS helps avoid false positives corresponding to partial masks from inter-frame matching and improves the performance on top of intra-frame and inter-frame matching. Our best results on both datasets are achieved by combining all three components. 

\begin{table}[t]
\centering
\small
\setlength{\tabcolsep}{1mm}
\vspace{-0.01\linewidth}
\begin{tabular}{|C{13mm}C{13mm}C{14mm}|C{13mm}C{13mm}|}
\hline
\multicolumn{3}{|c|}{MaskConsist Components}     & \multicolumn{2}{c|}{$AP_{50}$} \\ \hline
                            Intra-F & Inter-F & IoM-NMS & YTVIS     & Cityscapes    \\ \hline
            \xmark        &        \xmark          &    \xmark     & $31.43$     & $14.66$         \\ % \hline
         \xmark       & \cmark                & \cmark       & $33.75$     &     $14.92$           \\ % \hline
  \cmark   &   \xmark  & \cmark       & $31.08$     &       $14.43$        \\
 \cmark   & \cmark    &    \xmark     & $33.65$     &     $15.27$          \\ \hline
 \cmark             & \cmark                & \cmark       & $\textbf{34.66}$     & $\textbf{16.05}$         \\ \hline
\end{tabular}
\caption{Ablation study of MaskConsist components. The numbers in this table are generated by models with two-step training.}
\label{tab:maskconsist}
\vspace{-0.06\linewidth}
\end{table}

In Tab.~\ref{tab:mc}, we further explore the effectiveness of MaskConsist module by combining it with other weakly supervised instance segmentation methods: WISE~\cite{laradji2019masks} and IRN~\cite{ahn2019weakly}. Cross in the ``w/ MC'' column denotes the use of Mask R-CNN instead of MaskConsist. The results show that, by incorporating mask matching and consistency in the second stage of training, MaskConsist module consistently improves original weakly supervised methods by about $2\%$. Combining flowIRN module with MaskConsist achieves the best performance on both YTVIS and Cityscapes.

We also quantitatively evaluate how consistent the predictions of MaskConsist are on consecutive frames. As presented in the fifth column of Tab.~\ref{tab:mc}, we report the temporal consistency ($TC$) metric similar to \cite{liu2020efficient}. This metric measures the $AP_{50}$ between mask predictions and flow warped masks on consecutive frames in YTVIS. We observe consistent improvement in $TC$ by adding MaskConsist to training.

\begin{table}[t]
\centering
\vspace{-0.01\linewidth}
\small
\setlength{\tabcolsep}{1mm}
\begin{tabular}{|@{\hspace{1.2\tabcolsep}}L{15mm}|C{10mm}|C{15mm}|C{15mm}|C{15mm}|}
\hline
\multirow{2}{*}{Methods} & \multirow{2}{*}{w/ MC} &  \multicolumn{2}{c|}{$AP_{50}$} & \multirow{2}{*}{$TC$} \\ \cline{3-4} 
                         &                & YTVIS     & Cityscapes    &          \\ \hline
\multirow{2}{*}{WISE~\cite{laradji2019masks}}         &       \xmark          &   $24.54$   &   $10.51$  &    $72.08$     \\ 
         & \cmark         &   $27.03$    & $12.26$  &      $76.27$      \\ \hline
\multirow{2}{*}{IRN~\cite{ahn2019weakly}}          & \xmark       &   $29.64$      & $12.33$ &     $80.98$  \\ 
          & \cmark      &     $31.51$     & $14.72$ &   $82.04$    \\ \hline
\multirow{2}{*}{Ours}          & \xmark       &    $31.43$     &  $14.66$ &   $80.43$    \\ 
          & \cmark       &    $34.66$     & $16.05$  &     $84.36$  \\ \hline
\end{tabular}
\caption{MaskConsist works on top of different weakly supervised instance segmentation methods and improves both $AP_{50}$ and $TC$.}
\label{tab:mc}
\vspace{-0.03\linewidth}
\end{table}

\begin{figure}[t]
\centering
\vspace{-0.02\linewidth}
\includegraphics[width=\linewidth]{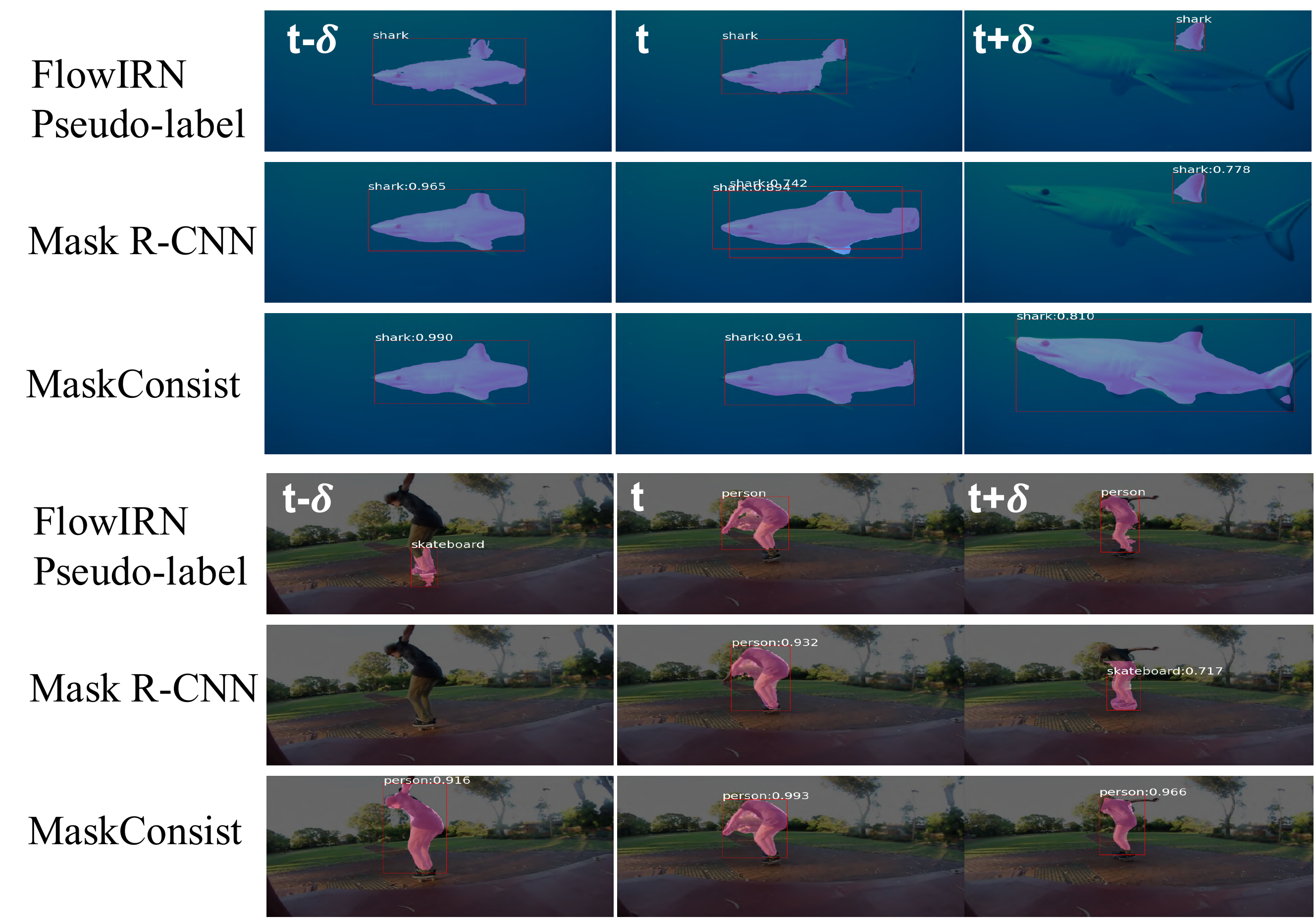}
\vspace{-0.06\linewidth}
\caption{Comparison of Mask R-CNN and MaskConsist on YTVIS. Both models are trained from flowIRN pseudo-label.}
\label{fig:maskconsist}
\vspace{-0.07\linewidth}
\end{figure}

In Fig.~\ref{fig:maskconsist}, we present two examples of Mask R-CNN and MaskConsist predictions on YTVIS clips. Both models are trained with flowIRN pseudo-labels. Mask R-CNN predictions are more susceptible to noisy pseudo-labels and less consistent across frames, while MaskConsist achieves more stable segmentation results.

\textbf{Further discussion:} Regarding the two types of errors presented in Fig.~\ref{fig:pull_figure}, we observe that our model has larger relative improvement over IRN on more strict metric: $27.3\%$ ($12.06\%$ vs. $9.47\%$) on $AP_{75}$, compared with $16.9\%$ ($34.66\%$ vs. $29.64\%$) on $AP_{50}$, indicating our model generates more accurate mask for high IoU metric. While our method outperforms IRN on $AP_{50}$, our method also predicts more instances per frame (avg. $1.81$ instances for ours vs. avg. $1.50$ instances for IRN), indicating our method is able to predict more instances with higher accuracy. These demonstrate that the two problems of partial segmentation and missing instance are both alleviated in our model.

%-------------------------------------------------------------------------
\vspace{-0.02\linewidth}
\section{Conclusion}
\vspace{-0.02\linewidth}
We observed that image-centric weakly supervised instance segmentation methods often segment an object instance partially or miss an instance completely. We proposed the use of temporal consistency between frames in a video to address these issues when training models from video frames. Our model (a) leveraged the constraint that pixels from the same instance move together, and (b) transferred temporally stable predictions to neighboring frames as pseudo-labels. We proved the efficacy of these two approaches through comprehensive experiments on two video datasets of different scenes. Our model outperformed the state-of-the-art approaches on both datasets.

\clearpage
\newpage
\onecolumn
\begin{appendices}
\section{Supplementary Material}
\subsection{Explanation of using optical flow gradient}
\begin{wrapfigure}{r}{0.32\textwidth}
\vspace{-0.25\linewidth}
  \begin{center}
    \includegraphics[width=0.3\textwidth]{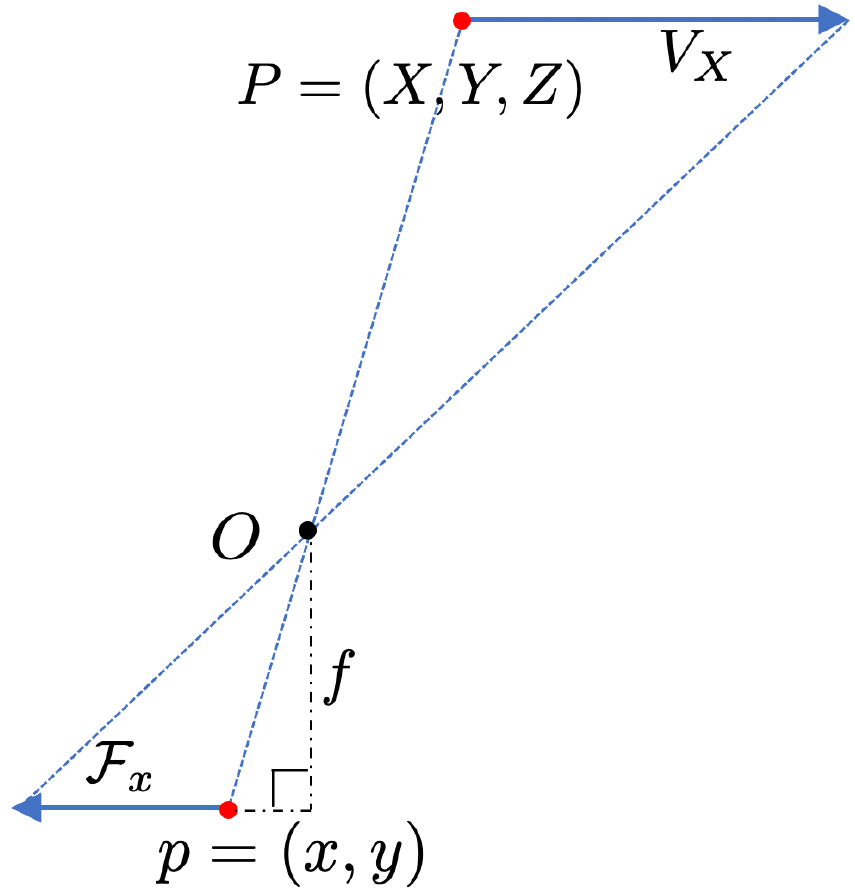}
  \end{center}
  \caption{3D motion results in 2D optical flow following imaging principle.}
  \label{fig:appflow}
  \vspace{-0.1\linewidth}
\end{wrapfigure}

Here we explain that spatial gradient of optical flow helps identify if two pixels are from the same instance.
Such information is encoded in the flow-boundary loss (Eq.~(3) in the manuscript). 

As shown in Fig.~\ref{fig:appflow}, let pixel $(x,y)$ in the image is a projection of point $(X,Y,Z)$ in the physical world with velocity $(V_X,V_Y,V_Z)$. The 3D motion results in optical flow on the image plane $\mathcal{F}_x, \mathcal{F}_y$. We consider for a short time window, most objects move in parallel to the image plane for Youtube VIS and Cityscapes data thus $V_z \approx 0$. For simple mathematical notation, we consider 3D motion only along X-axis $V_x=0$. The optical flow along image $x$ axis can be written as: 
\begin{align*}
\mathcal{F}_x = \frac{f*V_X}{Z}, 
\end{align*}
where $f$ is the camera focal length. We explain why we chose to use difference of optical flow first-order gradient instead of directly using difference of optical flow as followed.

For two neighboring pixels $p_i$ and $p_j$, if they are from the same rigid object, we have $V_{X_i}=V_{X_j} = V_{X}$, $d(V_X)/dx=0$. Then, their optical flow difference is:
\begin{align*}
\mathcal{F}_{x_i} - \mathcal{F}_{x_j} = &\frac{f*V_{X_i}}{Z_i} - \frac{f*V_{X_j}}{Z_j} = (\frac{1}{Z_i}-\frac{1}{Z_j})f*V_X, 
\end{align*}
which is not equal to zero when there is difference in depth for these two pixels. 

We propose to use the difference of spatial gradient of optical flow. The spatial gradient of flow along $x$ axis is defined as:
\begin{align*}
    \frac{d \mathcal{F}_x}{d x} = & \frac{d (1/Z)}{d x}f*V_X + \frac{d (V_X)}{d x}\frac{f}{Z} \\
    = & -\frac{1}{Z^2}\frac{d Z}{d x}f*V_X.
\end{align*}

For two neighboring pixels $p_i$ and $p_j$ on the same instance surface, assuming the surface is smooth thus $d Z_i/d x=d Z_j/d x=d Z/d x$, their difference of flow gradient is written as:
\begin{align*}
    \frac{d \mathcal{F}_{x_i}}{d x} -  \frac{d \mathcal{F}_{x_j}}{d x}
    = &-(\frac{1}{Z_i^2}-\frac{1}{Z_j^2})\frac{d Z}{d x}f*V_X
\end{align*}

In practice, the two pixels are in local neighbor and depth values $Z_i, Z_j$ are often large, thus $1/Z_i^2 -1/Z_j^2 \approx 0$ is better approximation than $1/Z_i -1/Z_j \approx 0$. This indicates using difference of optical flow gradient is a better signal for pixel affinity calculation.

A similar inference can also be applied to $y$ axis. In practice, we calculate the first-order gradient difference on both directions and encourage the norm to be zero.

\subsection{Prediction warping with optical flow}
In order to warp Mask R-CNN predictions with optical flow, we first warp the predicted masks using bi-linear interpolation as used in Spatial Transformer Network \cite{jaderberg2015spatial}. The warped mask is then converted to binary mask at threshold of $0.5$. Then the bounding box of warped prediction is generated from the warped mask and class label is directly copied.
In practice, we implement the mask warping function using \verb torch.nn.functional.grid_sample  in PyTorch~\cite{NEURIPS2019_9015} framework.  

\subsection{Training details for baseline methods}
\textbf{WISE~\cite{laradji2019masks}:} We train WISE using code published in \cite{PRMcode} and \cite{WISEcode}. For YTVIS, the model is trained for $20$ epochs with learning rate starting at $0.01$. For Cityscapes, the model is pre-trained on PASCAL VOC 2012 and then fine-tuned for $40$ epochs with learning rate starting at $0.001$. The MCG proposals are generated using code in \cite{MCGcode}. We generate pseudo-labels on the training split and train Mask R-CNN model as stated in Sec.~4.2 Implementation Details in the manuscript.

\textbf{F2F~\cite{lee2019frame}+MCG~\cite{pont2016multiscale}:} We first use F2F to generate semantic segmentation masks and then combine with MCG to generate instance masks as pseudo-labels. To train F2F on YTVIS and Cityscapes, we follow the F2F paper to generate the aggregated CAMs: warp the CAMs from $5$ consecutive frames to the key frame. The weakly supervised network backbone used in F2F is not available and we use SEC~\cite{kolesnikov2016seed} (code in \cite{SECcode}) which is advised by the F2F authors. For YTVIS, the model is trained for $10,000$ iterations with learning rate starting at $0.001$. For Cityscapes, the model is pre-trained on PASCAL VOC 2012 and then fine-tuned for $5,000$ epochs with learning rate starting at $0.001$. To combine with MCG proposals, we adopt a similar approaches in WISE\cite{laradji2019masks}. The resulting instance masks are used as pseudo-labels to train Mask R-CNN.

\textbf{IRN~\cite{ahn2019weakly}:} We train IRN using code published in \cite{IRNcode}. For YTVIS, the model is trained for $6$ epochs with learning rate starting at $0.1$. For Cityscapes, the model is pre-trained on PASCAL VOC 2012 and then fine-tuned for $6$ epochs with learning rate starting at $0.01$. The other training and inference parameters are set as default. The resulting instance masks on the training data are then used as pseudo-labels to train Mask R-CNN.

\textbf{IRN~\cite{ahn2019weakly}+F2F~\cite{lee2019frame}:} We use the flow-warped CAMs as in F2F to train IRN: warp CAMs from 5 neighboring frames to key frame to generate aggregated CAMs. The aggregated CAMs are then used to replace the CAMS used in original IRN. The training parameters are set the same as training original IRN model.

\subsection{More visualization of video intance segmentation results}
\begin{figure*}[t]
\centering
\includegraphics[width=\linewidth]{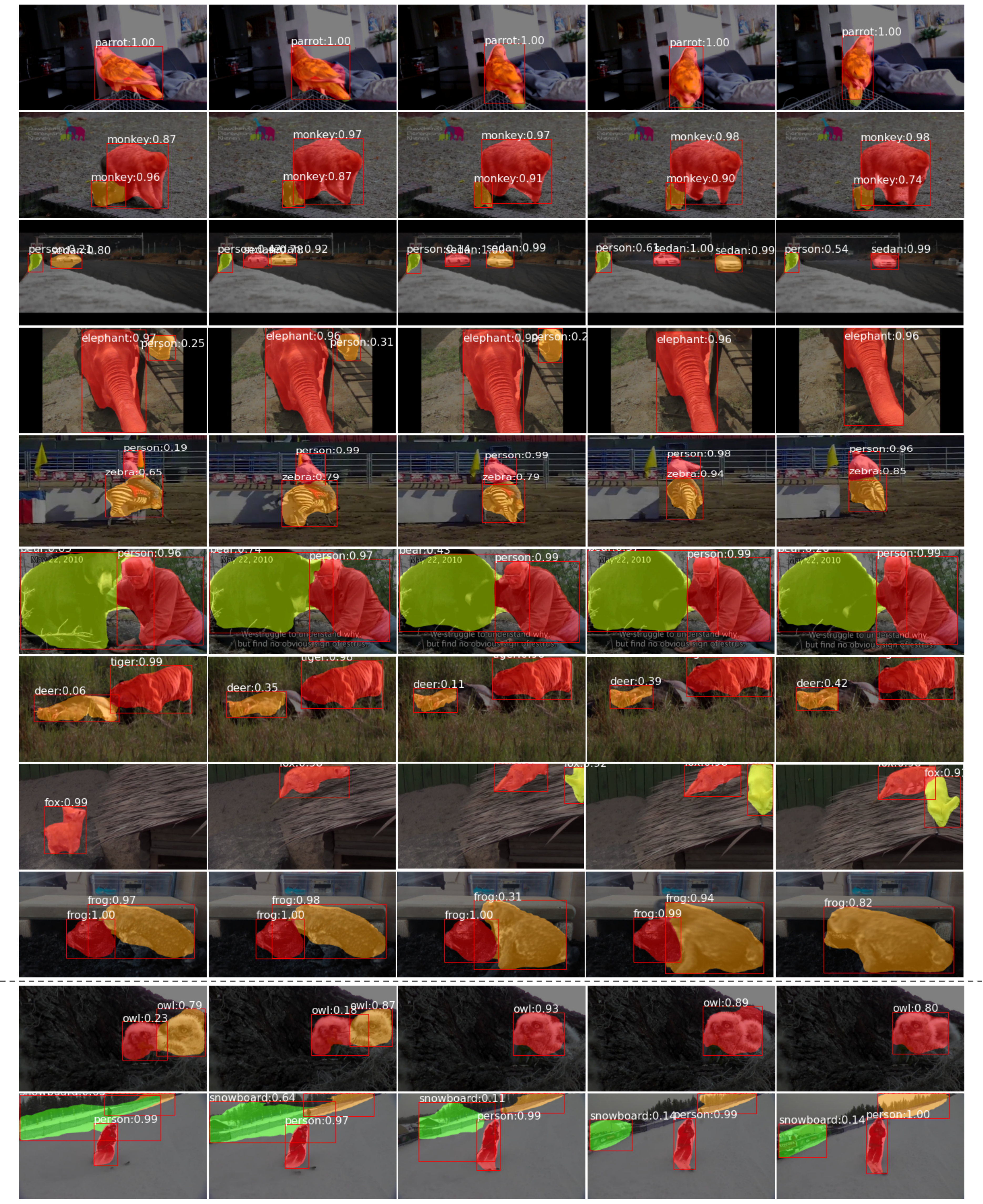}
\caption{More video instance segmentation results from our proposed method.}
\label{fig:app}
\end{figure*}

More video instance segmentation results are presented in Fig.~\ref{fig:app}. 
Every 5th frame in a video clip from YTVIS train\_val split are presented in each row. In the top nine examples, our method generates consistent instance masks with good object coverage and accurate instance boundary. We also include two failure cases in the last two rows, where the object is heavily occluded or object classes have strong co-occurrence pattern.

\end{appendices}
\twocolumn

\clearpage
\newpage
{\small
\bibliographystyle{ieee_fullname}
\bibliography{main}
}

\end{document}